\pdfoutput=1
\documentclass[11pt,twoside]{article} 
\usepackage{fancyheadings}
\usepackage{graphicx}    
\usepackage{xparse}

\newcommand*{\tensor}[1]{\overline{\overline{\boldsymbol{#1}}}}

\renewcommand*{\vec}[1]{\boldsymbol{#1}}

\usepackage{float}

\def\ba#1\ea{\begin{falign}#1\end{falign}}
\def\!#1\!{}

\def\bsa#1\esa{\begin{subequations}
\begin{gather}#1\end{gather} \end{subequations}}

\def\be#1\ee{\begin{equation}#1\end{equation}}

\newcommand{\Dp}[1]{\frac{d#1}{d p}}

\def\p{\partial}

\ExplSyntaxOn
\NewDocumentCommand{\mref}{m}{\quinn_mref:n {#1}}
\seq_new:N \l_quinn_mref_seq
\cs_new:Npn \quinn_mref:n #1
 {
  \seq_set_split:Nnn \l_quinn_mref_seq { , } { #1 }
  \seq_pop_right:NN \l_quinn_mref_seq \l_tmpa_tl
  ( 
  \seq_map_inline:Nn \l_quinn_mref_seq
    { \ref{##1},\nobreakspace } 
  \exp_args:NV \ref \l_tmpa_tl 
  ) 
 }
\ExplSyntaxOff
\usepackage{amsmath}
\usepackage{graphicx}
\usepackage{longtable}
\usepackage{caption}
\usepackage{subcaption}
\usepackage{amsmath}  
\usepackage{amssymb}
\usepackage{listings}
\usepackage{longtable}
\usepackage{setspace}
\usepackage{multirow}
\usepackage{array}
\usepackage{bm}
\usepackage{color}
\usepackage{soul}
\usepackage{bigints}
\usepackage{amstext}
\usepackage{float}
\usepackage{morefloats}
\usepackage{epstopdf}
\usepackage{bigints}
\usepackage[pagewise]{lineno}

\usepackage{amsfonts}
\usepackage{amsthm}
\usepackage{makecell}
\usepackage{array}
\newcolumntype{?}{!{\vrule width 2pt}}
\usepackage[pdftex,a4paper,colorlinks,linkcolor=blue,unicode=true, breaklinks=true,citecolor=blue]{hyperref}

\usepackage{algorithmic}
\usepackage{gensymb}
\usepackage{bigstrut}
\usepackage{rotating}
\usepackage{lmodern} 
\usepackage{tikz}
\usetikzlibrary{arrows}
\usepackage{booktabs}

\usepackage[sort]{natbib}
\usepackage{makeidx}
\usepackage{nomencl}
\RequirePackage{ifthen}
\renewcommand{\nomgroup}[1]{
\ifthenelse{\equal{#1}{A}}{\item[\textbf{Acronyms}]}{
\ifthenelse{\equal{#1}{B}}{\item[\textbf{Variables}]}{
\ifthenelse{\equal{#1}{C}}{\item[\textbf{Greek symbols}]}{
\ifthenelse{\equal{#1}{P}}{\item[\textbf{Superscripts}]}{
\ifthenelse{\equal{#1}{Q}}{\item[\textbf{Subscripts}]}
}}}}}
\makenomenclature

\usepackage{geometry}
\begin{document}
\newcommand\pix{Pix}
\newcommand\largepix{8cm}
\newcommand\medpix{4cm}
\newcommand\smallpix{3.5cm}
\newcolumntype{x}[1]{%
>{
  \centering\hspace{0pt}}p{#1}}%

\def\Dp#1{\dfrac{\partial #1}{\partial p}}
\def\DSp#1{\dfrac{\partial #1}{\partial S_{p'}}}
\def\DT#1{\dfrac{\partial #1}{\partial T}}

\def\Div{\mathop{\rm div}\nolimits}
\def\x{{\bf x}}
\def\y{{\bf y}}
\def\z{{\bf z}}
\def\A{{\bf A}}
\def\B{{\bf B}}
\def\R{{\bf R}}
\def\bg{{\bm \gamma}}
\def\dfrac#1#2{\displaystyle{\frac{#1}{#2}}}

\def\DelT[#1]{\textcolor{red}{\st{#1}}}
\def\DelTT[#1]{\textcolor{red}{#1}}
\def\CorT[#1]{\textcolor{blue}{\textbf{#1}}}
\def\CorTT[#1]{\textcolor{blue}{\textbf{#1}}}


\vspace{-5cm}

\renewcommand{\thefootnote}{\fnsymbol{footnote}}
\begin{center}

\noindent {\bf \LARGE 
A General Spatio-Temporal Clustering-Based\\ \vspace*{10pt} Non-local Formulation for Multiscale\\ \vspace*{16pt} Modeling of Compartmentalized Reservoirs} \\
\vspace*{20pt}

\noindent {Soheil Esmaeilzadeh\footnote{corresponding author: soes@stanford.edu}\\ Stanford University, CA, USA\\ \vspace*{10pt} Amir Salehi, Gill Hetz, Feyisayo Olalotiti-lawal, Hamed Darabi, David Castineira\\ Quantum Reservoir Impact LLC, TX, USA }
\end{center}

\section{Abstract}
Representing the reservoir as a network of discrete compartments with neighbor and non-neighbor connections is a fast, yet accurate method for analyzing oil and gas reservoirs. Automatic and rapid detection of coarse-scale compartments with distinct static and dynamic properties is an integral part of such high-level reservoir analysis. In this work, we present a hybrid framework specific to reservoir analysis for an automatic detection of clusters in space using spatial and temporal field data, coupled with a physics-based multiscale modeling approach. \\
A novel and rigorous non-local formulation for flow in porous media is presented, in which the reservoir is represented by an adjacency matrix describing the connectivities of comprising compartments. We automatically divide the reservoir into a number of distinct compartments, in which the direction-dependent multiphase flow communication is a function of non-local phase potential differences. Our proposed clustering framework begins with a mixed-type raw dataset which can be categorical/numerical, spatial/temporal, and discrete/continuous. The dataset can contain noisy/missing values of different data types including but not limited to well production/injection history, well location, well type, geological features, PVT measurements, perforation data, etc. Unsupervised clustering techniques suited to the input data types (e.g. k-prototypes, spectral, Gaussian Mixtures, and hierarchical clustering), and appropriate distance measures (such as Euclidean distance, soft dynamic time warping, and mode) are used. The input data is standardized, and upon convergence check, the best clustering representation is obtained. Finally, Support-Vector-Machine technique is utilized in the kernel space to trace a demarcating hyperplane for the clusters.\\
The proposed framework is successfully applied to more than five mature fields in the Middle East, South and North America, each with more than a thousand wells. In a specific case study reported here, the proposed workflow is applied to a major field with a couple of hundreds of wells with more than 40 years of production history. Leveraging the fast forward model, an efficient ensemble-based history matching framework is applied to reduce the uncertainty of the global reservoir parameters such as inter-blocks and aquifer-reservoir communications, fault transmissibilities, and block-based oil in place. The ensemble of history matched models are then used to provide a probabilistic forecast for different field development scenarios. In addition, the clustering framework enables us to treat missing data and use the augmented dataset for improving the clustering accuracy.\\
In summary, in this work a novel hybrid approach is presented in which we couple a physics-based non-local modeling framework with data-driven clustering techniques to provide a fast and accurate multiscale modeling of compartmentalized reservoirs. This research also adds to the literature by presenting a comprehensive work on spatio-temporal clustering for reservoir studies applications that well considers the clustering complexities, the intrinsic sparse and noisy nature of the data, and the interpretability of the outcome. \\ \\
\textbf{keywords}: Artificial Intelligence; Machine Learning; Spatio-Temporal Clustering; Physics-Based Data-Driven Formulation; Multiscale Modeling
\noindent
\section{Introduction}
In subsurface flow modeling, subsurface formations are characterized by the heterogeneity over multiple length scales, which can have a strong impact on the flow and transport \citep{Bonnet2001,Abu-Al-Saud2016,moatazaa,Cinar2014}. To account for this multiscale nature of the subsurface, high-resolution reservoir simulation is widely used to model the complex physics present in oil recovery processes. Although the recent increase in computational resources makes the flow simulation of large-scale models feasible, simulations with high-resolution models are still computationally expensive and prohibitive.\\
In the long term oil recovery processes, steps such as field development planning, resource allocation, and production optimization are of great importance. During these steps, reservoir simulation models are extensively used for objective function calculations and this can pose computational challenges as the forward models have to be run thousands of times \citep{Salehi2019a,Brown2017,Olalotiti2019,Shirangi2016}. \\
In reservoir analysis, calibrating techniques are used for estimating unknown properties of a reservoir such as porosity and permeability from historical measured data as a type of inverse problem where the historical data are usually taken at the production wells and might consist of pressure and/or flow data. The model is considered to be calibrated if it is able to reproduce the historical data of the reservoir it represents. This calibration process is called history matching, and this is the most time-consuming phase in any reservoir simulation study since it requires forward run of a fine-scale reservoir simulation model in an often exhaustive iterative process \citep{GharibShirangi2014,Tahmasebi2018}.\\
On the other hand, reservoir simulation models are subject to uncertainty in a great variety of input parameters caused by geological variability, measurement error, petrophysical-modeling uncertainty, and structural-modeling uncertainty, to name just a few. Furthermore, a correct analysis of uncertainty propagation through the model increases the quality and robustness of the decision-making process of field management. Consequently, an accurate analysis of uncertainty requires a large number of simulations \citep{Camacho2017,Lee2014a,Yeh2014,Yang2007}.  
Furthermore, the computational time of large-scale refined models becomes a bottleneck in the decision-making process and assimilating real-time data into a model \citep{salehi2013, Salehi2016,Gildin2011, Ghasemi2012}.\\
In all of the reservoir studies areas mentioned above, there have always been attempts to accelerate simulations in order to overcome their prohibitive costs and make them feasible for real-world applications while honoring the accuracy and reliability of the outcome. For example, using reduced order modeling (ROM), high-order reservoir models can be reorganized and reduced into (non-)linear low-order models using theoretical concepts like modal decomposition, balanced realization, proper orthogonal decomposition (POD), and system identification \citep{Bistrian2017,Fujimoto2010,Ghasemi2014,cardoso2010,Udy2017,Markovinvic2002}. ROM has received significant attention in the recent years particularly for use in optimization procedures in many areas of research, where a forward model must be run many times in order to determine the optimal design or the operating parameters \citep{Esmaeilzadeh2019a}. ROM procedures were first applied by \citet{Vermeulen2004} for subsurface flow problems. It has been observed that the standard POD procedures are not very robust and/or require more basis functions as the models become very nonlinear in the presence of strong gravitational effects with highly nonlinear relative permeability functions where the speed-up offered by standard POD techniques will lose its significance \citep{cardoso2009}. As another ROM technique, in the modal decomposition approach, the reduction that can be achieved is restricted by the low-order model and is not anymore able to reconstruct the behaviour of the high-order model accurately, as soon as more than half of the eigenmodes get truncated during the model reduction \citep{Heijn2004}. \\
On the other hand, in order to accelerate reservoir simulations, there have been attempts to decrease the number of grid cells in a simulation model by some coarsening techniques (upscaling methods) [\citep{Salehi2017}\citep{Salehi2019}]. Using homogenization during the coarsening process, a heterogeneous property region consisting of fine grid cells is replaced with an equivalent homogeneous region made up of a single coarse grid cell with an effective property value that needs to be calculated [see refs. \citep{Gautier1999,Fredrik1998,Durlofsky1996}]. A wide range of methods have been proposed in the literature to effectively carry out this coarsening (e.g. local, extended local, global, and quasi-global techniques for the upscaling of permeability, permeability upscaling in the vicinity of wells, flow-based grid generation techniques) which all try to increase the simulation speed while preserving a level of model accuracy [see ref. \citep{Wen1998,Holden2000,Durlofsky2003,Barker1997,Rabahy2002,Durlofsky1998}].\\
As an alternative approach to upscaling, a number of so-called multiscale methods have also been developed in order to accelerate the reservoir simulation process while achieving a higher level of accuracy. The key idea of the multiscale methods is to construct a set of operators that map between the unknowns associated with cells in a fine grid (holding the petrophysical properties of the geological reservoir model) and unknowns on a coarser grid used for dynamic simulation. The operators are computed numerically by solving localized flow problems in the same way as for flow-based upscaling methods. Unlike effective parameters, the multiscale basis functions have subscale resolution, which ensure that fine-scale heterogeneity is correctly accounted for in a systematic manner [see refs. \citep{Hajibeygi2012,Lee2008,Tchelepi2007,Darve2000,Forsum1978,Hajibeygi2009,Hajibeygi2011}] \\
As another way to accelerate reservoir simulations, \citet{patent:20150346010} proposed an approach where a reservoir is treated as a combination of multiple interconnected compartments which under a range of uncertainty can capture the reservoir's response during a recovery process. In this work, we extend the approach proposed by \citet{patent:20150346010} to represent a reservoir in a multiscale form consisting of multiple interconnected segments. To identify segments of the reservoir, we use the state-of-the-art spatial, temporal, and spatio-temporal unsupervised data-mining clustering techniques. Then, a novel non-local formulation for flow in porous media is presented, in which the reservoir is represented by an adjacency matrix describing the neighbor and non-neighbor connections of comprising compartments. We automatically divide the reservoir into $N$ distinct compartments, in which the direction-dependent multiphase flow communication is a function of non-local phase potential differences. Having the segmented reservoir and the uncertain parameters, we use a robust history matching technique to reproduce reservoir's historical response. Finally, for the segmented and history matched reservoir, we carry out forecasting that can be used in reservoir management and field development applications.\\
The manuscript is ordered as the following: First, we describe the methods and frameworks. We start by describing the spatio-temporal clustering framework, then we provide an overview of the compartmentalized reservoir simulation framework where we explain multi-tank history material balance and multi-tank predictive material balance. Finally, we describe the history matching approach. Afterwards, we present the results of clustering, multi-tank material balance, and history matching for a real-world reservoir.
\section{Methods and Frameworks}\label{sec_methods_frameworks}
\subsection{Clustering.} \label{sec_clustering} 
Clustering data-mining process is an interactive and iterative procedure that involves many steps, such as data extraction and cleaning, feature selection, algorithm design, and post-processing analysis of the output when an algorithm is applied to the dataset. The goal of clustering data-mining is to automate the discoveries of patterns and information, which can be examined by domain experts for further validation and verification. Clustering is one of the most fundamental tasks in spatial data-mining. It is the process of aggregating objects into different groups known as clusters such that the objects within the same cluster are similar to each other but dissimilar from those in other clusters. Clusters are generated on the basis of a \textit{similarity criterion}, which is used to determine the relationship between each pair of objects in the dataset. In this section, we present a general framework for clustering in subsurface applications as an unsupervised machine learning approach. We carry out clustering at the well-level and find separating hyperplanes of wells in the physical domain. The data can be either spatial or temporal, categorical or numerical. Well coordinates and PVT measurements are examples of spatial numerical data; well type, fault-regions, and well states are examples of spatial categorical data; and well bottomhole pressure, production/injection rates, cumulative production, gas/oil ratio, water/oil ratio are examples of temporal numerical data. \\
\subsection{Spatial Clustering.}\label{sec_spatial}
During the spatial clustering, we deal with features that are static (do not change over time) and can either be categorical (discrete categorical labels) or numerical (continuous real values). We build the feature vector using both categorical and numerical spatial characteristics of wells. The common approach for clustering numerical feature vectors is \textit{k-means}, which clusters numerical data based on \textit{Euclidean distance} \citep{Jain2010,Hartigan1979,Steinley2006,Alsabti1997}. For categorical data the \textit{k-modes} method is used, which defines clusters based on the number of matching categories between data points \citep{Huang1999,Huang2003,Chaturvedi2001}. However, for the case of having feature vectors comprised of mixed categorical and numerical data neither of the two approaches could be used. For this scenario, \textit{k-prototypes} first proposed by \citet{LiJie}, and \citet{Zhexue1998} got further studied later on by \citet{Ahmad2007} and \citet{Ji2012}.\\
In the k-prototypes approach, the cost function for clustering mixed datasets with $n$ data objects and $m$ attributes ($m_r$ numeric attributes, $m_c$ categorical attributes, $m = m_r + m_c$) is defined as:
\be\label{Eq_clust1}
\zeta = \sum_{i=1}^n D(d_i, C_j)\,,
\ee
where $D(d_i, C_j)$ is the distance of a data object $d_i$ from the closest cluster center $C_j$. $D(d_i, C_j)$ is defined as:
\be
D(d_i, C_j) = \sum_{t=1}^{m_r}(d_{it}^r - C_{jt}^r)^2 + \gamma_j\sum_{t=1}^{m_c} \delta (d_{it}^c, C_{jt}^c)\,,
\ee
where $d_{it}^r$ are values of numeric attributes and $d_{it}^c$ are values of categorical attributes for data object $d_i$. Here $C_j = (C_{j1}, C_{j2}, ..., C_{jm})$ represents the cluster center for cluster $j$. $C_{jt}^c$ represents the most common value (i.e. mode) for categorical attributes $t$ and class $j$.  $C_{jt}^r$ represents mean of numeric attribute $t$ and cluster $j$. For categorical attributes, $\delta(p,q) = 0$ when $p=q$ and $\delta(p,q) = 1$ when $p\neq q$. $\gamma_j$ is a weight for categorical attributes of cluster $j$. Cost function $\zeta$ is minimized for clustering mixed datasets. Equation \eqref{Eq_clust1} is a combined similarity measure on both numeric and categorical attributes between objects and cluster prototypes. The similarity measure for numeric attributes is the squared Euclidean distance, whereas the similarity measure for categorical attributes is the number of mismatches between cluster prototypes and objects. Weight $\gamma_j$ is introduced to avoid favoring either type of the attributes and is considered as a tuning hyperparameter. \\
The k-prototypes algorithm can be described as the following: 
(1). select $k$ initial prototypes from the dataset $\mathbf{X_{mn}}$, one for each cluster; (2). allocate each object in $\mathbf{X_{mn}}$ to a cluster whose prototype is the nearest to it according to the cost function in Equation \eqref{Eq_clust1} and update the prototype of the cluster after each allocation; (3). after all objects have been allocated to a cluster, recalculate the similarity of objects against the current prototypes. If an object is found such that its nearest prototype should be another cluster rather than its current one, reallocate the object to that cluster and update the prototypes of both clusters; (4). repeat (3) until no object has changed clusters after a full cycle test of $\mathbf{X_{mn}}$. The number of clusters here is an input parameter and is determined using the \textit{elbow criterion} \citep{Pearce1977,Yu2014} during the clustering process. Although, using other approaches such as density-based spatial clustering (aka DBSCAN) \citep{Borah2004} for finding out the value of $k$ a priori to clustering has been proposed, elbow criterion is reported to be a robust and independent approach \citep{Yan2005,Sugar2003}. 
\subsection{Adaptive Temporal Clustering.} \label{sec_temporal}
In this part, we explain our approach on how we carry out clustering in an adaptive way based on temporal (dynamically changing with time) properties including but not limited to quantities such as pressure profile, oil, gas, and water production rates, water and gas injection rates, gas/oil ratio, and water/oil ratio. We perform temporal clustering as explained in the following of this part. After carrying out temporal clustering, for cluster adaptiveness, we examine each temporally clustered group of wells and if the internal variation of the signals is larger than a pre-specified threshold, we sub-cluster those clusters again. Our approach can easily be used for any temporal quantity in a similar scenario as explained below. \\
There have been different approaches proposed for temporal proximity-based clustering using similarity measures such as Euclidean distance, Pearson correlation coefficient, Kullback-Liebler (KL) divergence, and dynamic time warping (DTW), and using clustering algorithms such as k-means, k-medoids, hierarchical, and kernel DBScan \citep{Aghabozorgi2015,Rani2012}. In this work, we use \textit{dynamic time warping (DTW)} as the \textit{similarity measure} of our time series due to its advantages over other similarity counterparts such as being able to give the position of alignment between two sequences, and being able to compare sequences with different lengths. DTW also can exploit the temporal distortions and compare shifted or distorted evolution profiles whose time sampling can even be irregular due to its optimal alignment. DTW is also able to find optimal global alignment between sequences and accordingly is the most commonly used measure to quantify the dissimilarity between sequences \citep{Kruskall1983,Aach2001,Bar-Joseph2002,Gavrila1995,Rath}. DTW also provides an overall real number that quantifies the similarity between two temporal sequences and makes it possible to find the best global alignment between them. As the clustering approach, here for the illustration purposes, we have implemented and used \textit{k-means} together with DTW as the similarity measure, however, any other clustering approach can be used and evaluated in a similarly consistent way. \\
DTW was first introduced by \citet{SakoeChiba71} and \citet{Hiroaki1978}, with applications to speech recognition. It finds the optimal alignment (or coupling) between two sequences of numerical values and captures flexible similarities by aligning the coordinates inside both sequences. \\
In the following, we briefly describe how DTW works on two given sequences.
Given two sequences $X=\lbrace x_1,x_2,...,x_n\rbrace$ and $ Y = \lbrace y_1,y_2,...,y_m\rbrace$ with different lengths ($N\neq M$), a warping path $W$ is an alignment between $X$ and $Y$, involving \textit{one-to-many} mapping for each pair of elements. The \textit{cost} of a warping path is calculated by the sum of each mapping pair cost. Furthermore, the warping path contains 3 constraints: (1). \textit{Endpoint constraint,} which states that the alignment starts at pair $(1, 1)$ and ends at pair $(N, M)$; (2). \textit{Monotonicity constraint,} which states that the order of elements in the path for both $X$ and $Y$ should be preserved same as the original order in $X$ and $Y$, respectively; (3). \textit{Step size constraint,} which states that the difference of index for both $X$ and $Y$ between two adjacent pairs in the path need to be \textit{no more than} one step (i.e. $(x_i, y_j)$ pair can be followed by three possible pairs including $(x_{i+1}, y_j ), (x_i, y_{j+1})$, and
$(x_{i+1}, y_{j+1})$). As DTW is a distance measure that
searches the optimal warping path between two series, in DTW we firstly construct a cost matrix $C$ where each element $C(i, j)$ is a cost of the pair $(x_i, y_j )$ specified by using Euclidean, Manhattan, or other distance functions. The initial step of DTW algorithm is defined as:
\begin{equation} 
DTW(i,j) = 
     \begin{cases}
       \infty & \text{if } (i=0 \text{ or } j=0) \text{ and } i \neq j\\
       0 & \text{if } i=j=0
     \end{cases}
\end{equation}
The recursive function of DTW is defined as:
\begin{equation}
DTW(i,j) = 
     \begin{cases}
       DTW(i-1,j) + \omega_h\,C(i,j)\\
        DTW(i,j-1) + \omega_v\,C(i,j)\\
        DTW(i-1,j-1) + \omega_d\,C(i,j)
     \end{cases}
\end{equation}
where $\omega_h, \omega_v, \omega_d$ are weights for horizontal, vertical, and diagonal directions, respectively. $DTW(i,j)$ denotes distance or
cost between two subsequences $(x_1, ..., x_i)$ and $(y_1, ..., y_j)$,
and $DTW(N, M)$ indicates the total cost of the optimal warping
path. DTW is calculated based on dynamic programming of the order \textbf{O}($M\times N$). In an equally weighted case, $(\omega_h, \omega_v, \omega_d) = (1, 1, 1)$, the recursive function has the preference for diagonal alignment direction because the diagonal alignment takes one-step cost while the combination of a vertical and a horizontal alignment takes two-steps cost where this preference can be counterbalanced if needed by, for instance, setting $(\omega_h, \omega_v, \omega_d) = (1, 1, 2)$. One potential issue of using this DTW definition is that the longer the two sequences are, the larger their DTW value will be, so its absolute value may not truly reflect the difference of the two sequences. Thus, we use the normalized DTW, defined by dividing the original DTW by the sum of the lengths of two sequences as:
\begin{equation}
DTW_{norm}(N,M) = \frac{DTW(N,M)}{N+M}\,.
\end{equation}
Each alignment in the warping path has a corresponding weight, selected from $(\omega_h, \omega_v, \omega_d)$ and the sum of the weights for all alignments equals to the sum of the lengths of two sequences i.e. $N + M$. Therefore, the normalized DTW evaluates the average distance of alignments in the warping path for two sequences. 
\subsection{Spatio-temporal Clustering.} \label{sec_spatioTemporal}
Once the quantities of interest for clustering are chosen, depending on the type of features, we use \textit{spatial clustering} and/or \textit{temporal clustering} as presented before for clustering a given field at the well-level. In the next step, we want to divide a given reservoir into separate zones in the physical space based on the well-level clusters obtained. There have been similar attempts in a few previous works as so-called \textit{spatially constrained multivariate clustering} \citep{Coppi2010,ChatzisSotiriosP.2008,Legendre1987,Patil2006}, however, these attempts in enforcing spatial coherence as constraints in the clustering framework have not been successful in the cases with multiple interrelated mixed type features. Also, in the spatially constrained multivariate clustering approaches, defining the proper constraints and satisfying them at the cluster level is a challenging step. In this work, we propose an alternative approach that is not only robust while dealing with mixed type temporal and spatial features but also easy to implement and tune. \\
Once the clusters at the well-level have been formed using spatial and/or temporal clustering, we find the separating hyperplanes between well-level clusters in order to divide a given field into separate zones. We shape this problem as a supervised learning and use \textit{support vector machine (SVM)} approach as a discriminant classifier \citep{Suykens1999,Scholkopf1998}. We consider wells as individual samples, the cluster labels of the wells as categorical target variables, and the wells' head coordinates as numerical features. Forming the problem this way, SVM as a classifier predicts the separating hyperplanes of the clustered wells. To account for separations in higher feature space using the \textit{kernel trick} we consider different kernel functions such as linear polynomial and exponential, and also account for regularization to avoid overfitting. For the details on support vector machines see \citep{Scholkopf:2001:LKS:559923,9783540243885,Lin2012}.
\subsection{Compartmentalized Reservoir Modeling.} \label{sec_history_forecast} In the following sections, we explain our proposed model to study a compartmentalized reservoir with a non-local formulation where the reservoir is represented by an adjacency matrix describing the neighbor and non-neighbor connections of the comprising compartments. We will look into the reservoir's history using a so called material balance network (MatNet) methodology and also propose a forecast approach using a predictive material balance network.
\subsection{Reservoir's History - Material Balance Network (MatNet).}
\label{Sec_MBal_History}
In this section, we present the generalized material balance equation that relies on the following assumptions: (1). the reservoir is under an isothermal transformation and is at equilibrium at any point in time; (2). gas component can be dissolved in the oil phase; (3). oil component can be volatile in the gas phase; (4). and oil, water, and rock are slightly compressible.
We start by expressing the initial volume of gas component in the reservoir ($G$) as sum of the volumes present as the gas component in the gas phase and as the dissolved gas in the oil phase at the reservoir conditions. Similarly, the original oil in place ($N$) can be expressed as sum of the volumes present as the oil component in the oil phase and as the volatile oil in the gas phase at the reservoir conditions as:
\bsa
G = G_{fgi} + N_{foi}R_{si}\,, \label{Eq_GN11}\\ 
 N = N_{foi} + G_{fgi}R_{vi}\,. \label{Eq_GN12}
\esa
After some time of recovery from the reservoir, the remaining oil/gas in place can be expressed as the difference between the initial volume of oil/gas in place and the cumulative volume of oil/gas produced as:
\bsa 
G - G_p = G_{fg} + N_{fo}R_s\,, \label{Eq_GN21} \\ 
N - N_p = N_{fo}+G_{fg}R_{v}\,. \label{Eq_GN22}
\esa
Using Equations \mref{Eq_GN11, Eq_GN12,Eq_GN21,Eq_GN22} we get the volume of gas component in the gas phase ($G_{fg}$) and the volume of oil component in the oil phase ($N_{fo}$) as:
\bsa 
G_{fg} =  \frac{G_{fgi} + N_{foi}R_{si} - G_p-(N_{foi} + G_{fgi}R_{vi}-N_p)\,R_s}{1-R_sR_v}\,, \label{Eq_gfg}\\
N_{fo} = \frac{N_{foi} + G_{fgi}R_{vi} - N_p-(G_{fgi} + N_{foi}R_{si}-G_p)\,R_v}{1-R_sR_v}\,. \label{Eq_nfo}
\esa
We use the original reservoir volume as a control volume to write the material balance equation as:
\bsa \label{Eq_Summation_orig}
\Delta V_{local} + \Delta V_{non-local} = 0\,,\\
\Delta V_{local} =  \Delta V_o + \Delta V_g + \Delta V_w + \Delta V_r\,, \\
\Delta V_{non-local}  = \Delta V_{Neigh}\,,
\esa
where $\Delta V_{local}$ is the local volume change within a given block of the reservoir due to internal changes in pressure or local injections and productions leading to changes in oil, gas, water, and rock volumes, and $\Delta V_{non-local}$ is the change of volume due to non-local volumetric fluxes of different phases from other blocks of the reservoir (i.e. neighbor or non-neighbor connected blocks). $\Delta V$ here represents the difference between the initial and the current volumes at the reservoir conditions. The current volume of oil, gas, and water at reservoir conditions can be written as:
\bsa \label{Eq_GN5}
V_o = B_o\,(N_{foi} + G_{fgi}R_{vi}-G_{fg}R_v - N_p)\,,\\
V_g = B_g\,(G_{fgi} + N_{foi}R_{si} - N_{fo}R_s - G_p ) + G_{inj}B_{ginj}\,,\\
V_w = W_e - B_w W_p + B_{winj}W_{inj} + V_{\phi i}S_{wi}\,(1+c_w\Delta p)\,,
\esa
with the original volume of oil, gas, and water at the reservoir conditions being $V_{oi} =  N_{foi} B_{oi}$, $V_{gi} = G_{fgi}B_{gi}$, and $V_{wi} = V_{\phi i } S_{wi}$, respectively. The difference between the initial and the current oil/gas/water volumes at the reservoir conditions becomes:
\bsa
 \label{Eq_Vo}
\Delta V_o = V_{oi} - V_o = N_{foi}B_{oi} - B_o \,(N_{foi} + G_{fgi}R_{vi} - G_{fg}R_v - N_p )\,,\\
\label{Eq_Vg}
\Delta V_g = V_{gi} - V_g = G_{fgi}B_{gi} - G_{inj}B_{ginj} - B_g\,(G_{fgi} + N_{foi}R_{si} - N_{fo}R_s - G_p )\,,\\
\label{Eq_Vw}
\Delta V_w = V_{wi} - V_w = - W_e + B_w W_p - B_{winj}W_{inj} - V_{\phi i}S_{wi}c_w\Delta p\,,
\esa
with $V_{\phi i }$, $c_w$, and $\Delta p$ being the initial pore volume, the water compressibility, and the change of reservoir pressure, respectively.
Finally, the pressure depletion applied to the rock phase creates a shrinkage of the reservoir pore volume ($\Delta V_r$) as:
\be\label{Eq_Vr}
\Delta V_r = -V_{\phi i} c_f \Delta p\,,
\ee
where $c_f$ represents the formation (rock) compressibility. Finally, the volume change due to contributions from adjacent compartments (i.e. neighboring communications) can be expressed as: 
\be\label{Eq_Neigh}
 \Delta V_{Neigh} = \sum_{j=1}^{Neigh} \sum_{\alpha\,=\,o,g,w} \int_0^t \frac{T_{j}}{\bar{\mu}^t_{\alpha}}k_{r\alpha}(S^t)[p_{j,\alpha}^t - p_{\alpha}^t - \bar{\rho}_\alpha^tg(z_j - z)]\,dt\,.
\ee
By substituting Equations \mref{Eq_Vo,Eq_Vg,Eq_Vw,Eq_Vr,Eq_Neigh} in Equation \eqref{Eq_Summation_orig} and expressing the initial pore volume as a function of the initial reservoir fluid volumes as $V_{\phi i} = (N_{foi}B_{oi} + G_{fgi}B_{gi})/(1-S_{wi})$ we get:
\be \label{Eq_f1}
\begin{gathered}
[N_{foi}B_{oi} - (N_{foi} + G_{fgi}R_{vi} - G_{fg}R_v - N_p)B_o] -
[(N_{foi}B_{oi} + G_{fgi}B_{gi})\,\frac{c_f + c_wS_{wi}}{1-S_{wi}}\Delta p]\,+\, \\
[G_{fgi}B_{gi}-(G_{fgi} + N_{foi}R_{si} - N_{fo}R_s - G_p)B_g - G_{inj}B_{ginj}]\,+\, 
[B_wW_p - B_{winj}W_{inj} - W_e] + \\
\sum_{j=1}^{Neigh} \sum_{\alpha\,=\,o,g,w} \int_0^t \frac{T_{j}}{\bar{\mu}^t_{\alpha}}k_{r\alpha}(S^t)[p_{j,\alpha}^t - p_{\alpha}^t- \bar{\rho}_\alpha^tg(z_j - z)]\,dt  = 0\,.
\end{gathered}
\ee
Finally, using Equations \eqref{Eq_gfg} and \eqref{Eq_nfo} in Equation \eqref{Eq_f1} and by some rearrangements and simplifications we get the general material balance equation as:
\be \label{Eq_MBAL1}
\begin{gathered} 
N_{foi}\,\bigg[  \frac{B_o -B_{oi} + B_g(R_{si} - R_s) + R_v(B_{oi}R_s - B_oR_{si})}{1-R_sR_v}\bigg] -N_p \frac{B_o - R_sB_g}{1 - R_sR_v} - G_p \frac{B_g - R_vB_o}{1- R_sR_v} \,+\,\\
G_{fgi}\,\bigg[  \frac{B_g -B_{gi} + B_o(R_{vi} - R_v) + R_s(B_{gi}R_v - B_gR_{vi})}{1-R_sR_v}\bigg] -W_pB_w + W_{inj}B_{winj} + G_{inj}B_{ginj} \,+\, \\
(N_{foi}B_{oi} + G_{fgi}B_{gi})\frac{c_f + c_wS_{wi}}{1- S_{wi}} \Delta p + W_e \, + 
\sum_{j=1}^{Neigh} \sum_{\alpha\,=\,o,g,w} \int_0^t \frac{T_{j}}{\bar{\mu}^t_{\alpha}}k_{r\alpha}(S^t)[p_{j,\alpha}^t - p_{\alpha}^t- \bar{\rho}_\alpha^tg(z_j - z)]\,dt = 0\,.
\end{gathered}
\ee
In order to solve the Equation \eqref{Eq_MBAL1} as a coupled non-linear system of equations for all the compartments, we write it in the Residual form and after building the Jacobian matrix, we solve for the unknown pressure of each compartment as:
\bsa
 \label{Eq_Res}
 \vec{p}^{\,n} = [{p}_1^{\,n}, ..., {p}_N^{\,n}] = [{p}^{\,n}_{b}] \quad \text{with block index:~ } b = 1, ...\,, N, \\
\vec{R}(\vec{p}^{\,n}) = \vec{0}\,,\\
\label{Eq_Jac}
\tensor{J}\vec{\delta_p} = -\vec{R} \quad \text{where~~} J_{ij} = \frac{\partial{R}_i}{\partial{p}_j}, \quad \text{with } 1 \leq i,j \leq N\,,\\
\vec{p}^{\,n+1} = \vec{p}^n + \vec{\delta_p}\,.
\esa
For an arbitrary block $i$, we split the Residual into two parts of local and non-local, according to the Equation \eqref{Eq_Summation_orig} as $R_i = R_{i, local} + R_{i, non-local}$ where the local and non-local Residuals, respectively, are:
\begin{equation} \label{Eq_ResLoc1}
\begin{gathered} 
R_{i, local} = N_{foi}\,\bigg[ \frac{B_o -B_{oi} + B_g(R_{si} - R_s) + R_v(B_{oi}R_s - B_oR_{si})}{1-R_sR_v}\bigg] -N_p \frac{B_o - R_sB_g}{1 - R_sR_v} - G_p \frac{B_g - R_vB_o}{1- R_sR_v} \,+\,\\
G_{fgi}\,\bigg[ \frac{B_g -B_{gi} + B_o(R_{vi} - R_v) + R_s(B_{gi}R_v - B_gR_{vi})}{1-R_sR_v}\bigg] -W_pB_w + W_{inj}B_{winj} + G_{inj}B_{ginj} \,+\, \\
(N_{foi}B_{oi} + G_{fgi}B_{gi})\frac{c_f + c_wS_{wi}}{1- S_{wi}} \Delta p + W_e \,,
\end{gathered}
\end{equation}
\begin{equation} \label{Eq_ResLoc2}
\begin{gathered} 
R_{i, non-local} = 
\sum_{j=1}^{Neigh} \sum_{\alpha\,=\,o,g,w} \int_0^t \frac{T_{ij}}{\bar{\mu}^t_{\alpha}}k_{r\alpha}(S^t)[p_{j,\alpha}^t - p_{i,\alpha}^t - \bar{\rho}_\alpha^tg(z_j - z)]\,dt\,.
\end{gathered}
\end{equation}
To calculate the Jacobian matrix in Equation \eqref{Eq_Jac}, we split the derivative of the Residual of an arbitrary block $i$ into two parts, one with respect to the compartment $i$ and one with respect to the connected compartment $j$ to $i$, as:
\begin{equation} \label{Eq_Jacob1}
    J_{ij}  =
\left\{
	\begin{array}{ll}
	 \frac{\partial{R}_i}{\partial{p}_i} = \frac{\p R_{i,local}}{\p p_i} + \frac{\p R_{i,Neigh}}{\p p_i}  & i = j\\
      \frac{\partial{R}_i}{\partial{p}_j} = \frac{\p R_{i,Neigh}}{\p p_j} & i \neq j
	\end{array}\, ,
\right.
\end{equation}
where the derivative of the Residual terms in Equation \eqref{Eq_Jacob1} can be calculated, respectively, as:
\be\label{Eq_jacob1}
\begin{gathered} 
\frac{\p R_{i,local}}{\p p_i} = -W_p\Dp{B_w} + W_{inj}\Dp{B_{winj}} + G_{inj}\Dp{B_{ginj}} + \Dp{W_e} \,+ \,\\
N_{foi}\,\bigg\{\frac{\Dp{B_o}(1-R_vR_{si}) + \Dp{B_g}(R_{si}-R_s) + \Dp{R_s}(R_vB_{oi}-B_g) + \Dp{R_v}(B_{oi}R_s-B_oR_{si})}{1-R_sR_v}\bigg\} \,+\, \\
N_{foi}\,\bigg\{ \frac{\big[R_s\Dp{R_v} + R_v\Dp{R_S}\big]\big[B_o - B_{oi} + B_g(R_{si}-R_s) + R_v(B_{oi}R_s - B_oR_{si})\big]}{(1-R_sR_v)^2}\bigg\} \,+\, \\
G_{fgi}\,\bigg\{\frac{\Dp{B_g}(1-R_sR_{vi}) + \Dp{B_o}(R_{vi}-R_v) + \Dp{R_v}(R_sB_{gi}-B_o) + \Dp{R_s}(B_{gi}R_v-B_gR_{vi})}{1-R_sR_v}\bigg\} \,+\, \\
G_{fgi}\,\bigg\{ \frac{\big[R_s\Dp{R_v} + R_v\Dp{R_S}\big]\big[B_g - B_{gi} + B_o(R_{vi}-R_v) + R_s(B_{gi}R_v - B_gR_{vi})\big]}{(1-R_sR_v)^2}\bigg\} \,-\, \\
N_p \,\bigg\{\frac{\big[\Dp{B_o} - R_s\Dp{B_g} \ B_g\Dp{R_s} \big](1- R_sR_v) + \big[R_s \Dp{R_v} + R_v \Dp{R_s} \big](B_o - R_sB_g)}{(1-R_sR_v)^2} \bigg\}\,-\,\\
G_p \,\bigg\{\frac{\big[\Dp{B_g} - R_v\Dp{B_o} \ B_o\Dp{R_v} \big](1- R_sR_v) + \big[R_s \Dp{R_v} + R_v \Dp{R_s} \big](B_g - R_vB_o)}{(1-R_sR_v)^2} \bigg\}\,-\,\\
(N_{foi}B_{oi} + G_{fgi}B_{gi})\frac{c_f + c_wS_{wi}}{1- S_{wi}}\,,
\end{gathered}
\ee
\be\label{Eq_jacob2}
\frac{\partial {R_{i,\,Neigh}}}{\partial {p}^{\,n+1}_i} = 
\sum_{j=1}^{Neigh}\sum_{\alpha\,=\,o,g,w} T_{ij}\,k_{r\alpha}(S^n)\,\Delta t^{n+1}  \Big\lbrace\frac{1}{\bar{\mu}^{n+1}_{\alpha}}\Big[-1-g(z_j - z)\frac{\p \bar{\rho}_\alpha^{n+1}}{\p p^{n+1}_i}\Big] + \Big[p^{n+1}_j - p^{n+1}_i - \bar{\rho}_\alpha^{n+1}g(z_j - z)\Big]\frac{\p}{\p p^{n+1}_i}(\frac{1}{\bar{\mu}^{n+1}_{\alpha}})\Big\rbrace\,,
\ee
\be \label{Eq_jacob3}
\frac{\partial {R_{i,\,Neigh}}}{\partial {p}^{\,n+1}_j} = 
\sum_{\alpha\,=\,o,g,w} T_{ij}\,k_{r\alpha}(S^n)\,\Delta t^{n+1} \Big\lbrace{\frac{+1}{\bar{\mu}^{n+1}_{\alpha}}\Big[1-g(z_j - z)\frac{\p \bar{\rho}_\alpha^{n+1}}{\p p^{n+1}_j}\Big] + \Big[p^{n+1}_j - p^{n+1}_i - \bar{\rho}_\alpha^{n+1}g(z_j - z)\Big]\frac{\p}{\p p^{n+1}_j}(\frac{1}{\bar{\mu}^{n+1}_{\alpha}})}\Big\rbrace\,,
\ee
where Equations \mref{Eq_jacob1,Eq_jacob2} and \eqref{Eq_jacob3} are, respectively, the diagonal and off-diagonal terms in the Jacobian matrix.\\
As the aquifer model, we use \citet{Fetkovich1971} and \citet{Tarek}'s approach to define the cumulative volume of water encroached from the aquifer as a function of the reservoir's pressure drop where the cumulative encroached water volume $W^n_e$ is computed recursively at each time step $n$ as:
\be \label{Eq_fed0}
\Delta W^n_e = (W^n_e - W^{n-1}_e) = \frac{W_{ei}}{p_i}\Big[p_i(1 - \frac{W^{n-1}_e}{W_{ei}}) - \frac{1}{2}(p^n + p^{n-1}) \Big] \Big[1-\exp(-\frac{Jp_i\Delta t^n}{W_{ei}}) \Big]\,,
\ee
where $W_{ei} = c_t W_i p_i(\frac{\theta}{360})$ with $c_t$ being the total aquifer's compressibility as $c_t = c_f + c_w$. Rearranging Equation \eqref{Eq_fed0}, we get the aquifer's cumulative encroached water volume at each time step $n$ as:
\be \label{Eq_Fed1}
W^n_e = W^{n-1}_e + \frac{W_{ei}}{p_i}\Big[p_i(1 - \frac{W^{n-1}_e}{W_{ei}}) - \frac{1}{2}(p^n + p^{n-1}) \Big] \Big[1-\exp(-\frac{Jp_i\Delta t^n}{W_{ei}}) \Big]\,.
\ee
Assuming that $\frac{p_n + p_{n-1}}{2} \approx p_n$, Equation \eqref{Eq_Fed1} can be written as:
\be \label{Eq_fed2}
W^n_e = e^{-\frac{Jp_i\Delta t^n}{W_{ei}}}\,W_e^{n-1} + \frac{W_{ei}}{p_i}(p_i - p^n)(1 - e^{-\frac{Jp_i\Delta t^n}{W_{ei}}})\,.
\ee
We can rewrite Equation \eqref{Eq_fed2} as: 
\be \label{Eq_fed3}
W^n_e = a_nW^{n-1}_e + b_n\,,
\ee
where $a_n = e^{-\frac{Jp_i\Delta t^n}{W_{ei}}}$ and $b_n = \frac{W_{ei}}{p_i}(p_i - p^n)(1-e^{-\frac{Jp_i\Delta t^n}{W_{ei}}})$. For a few time steps, Equation \eqref{Eq_fed3} can be written as: 
\bsa
 \label{Eq_steps_we}
W_e^1 = a_1W_e^0 + b_1 = b_1\,,\\
W_e^2 = a_2W_e^1 + b_2 = b_1a_2 + b_2\,,\\
W_e^3 = a_3W_e^2 + b_3 = b_1a_2a_3 + b_2a_3 + b_3\,,\\
\vdots \nonumber\\
W_e^n = \sum_{k=1}^{n-1}b_k\Big( \prod_{j=k+1}^na_j \Big) + b_n\,.
\esa
As $\displaystyle\prod_{j=k+1}^na_j = \prod_{j=k+1}^n \exp(-\frac{Jp_i\Delta t_j^n}{W_{ei}}) = \exp(-\frac{Jp_i(t^n - t^k)}{W_{ei}})$, $W_e^n$ can be written as: 
\be \label{Eq_wen2}
W_e^n = \sum_{k=1}^n \frac{W_{ei}}{p_i}\Big[1-e^{(-\frac{Jp_i\Delta t^{\,k}}{W_{ei}})}\Big]e^{-\frac{Jp_i(t^{\,n} - t^{\,k})}{W_{ei}}}(p_i - p_k)\,.
\ee
The derivative of $W_e$ with respect to reservoir's pressure can also be calculated as:

\begin{equation} \label{Eq_we_jacob}
\frac{\partial W_e^k}{p_j} = 
     \begin{cases}
       \frac{-W_{ei}}{p_i}\Big[1-e^{(-\frac{Jp_i\Delta t^{j}}{W_{ei}})}\Big]e^{-\frac{Jp_i(t^{\,k} - t^{\,j})}{W_{ei}}}& j\leq k \\
       0 & j > k
     \end{cases}
\end{equation}

Finally, using the Fetkovich's approach, the aquifer terms in the Residual and Jacobian Equations \mref{Eq_ResLoc1,Eq_jacob1} can be calculated using the above Equations \mref{Eq_wen2, Eq_we_jacob}.
\subsection{Reservoir's Forecast - Predictive Material Balance Network.} \label{sec_forecast}
The material balance network approach presented in the previous section is used for analysis of the reservoir based on the available historical data. In this section, we extend the formulation to forecast the reservoir's performance based on the historical data and the planned schedule of recovery in the future. In particular, a new algorithm is developed for three-phase flows in a mixed-drive reservoir, capable of predicting flow rates and pressure, assuming pressure constraints on the producing wells. Existing methods in the literature by \citet{Tarner}, \citet{Muskat1945}, and \citet{Tracy1995} are designed for solution-gas drive reservoirs with two-phase flows. In the historical material balance network analysis, the flow rates corresponding to each phase are available based on the historical data, however, in the predictive mode, the rates are forecasted and therefore, the constraint on the producer wells is of flowing bottomhole pressure type. \\ 
At each time step, there are four unknowns as pressure ($p$), cumulative oil production ($N_p$), cumulative gas production ($G_p$), and cumulative water production ($W_p$). Therefore, four equations are required at each time step to solve for the four unknowns.\\
The first equation is the general material balance equation \eqref{Eq_MBAL1} as:
\be \label{Eq_forecast1}
\begin{gathered} 
N_{foi}\,\bigg[  \frac{B_o -B_{oi} + B_g(R_{si} - R_s) + R_v(B_{oi}R_s - B_oR_{si})}{1-R_sR_v}\bigg] -N_p \frac{B_o - R_sB_g}{1 - R_sR_v} - G_p \frac{B_g - R_vB_o}{1- R_sR_v} \,+\,\\
G_{fgi}\,\bigg[  \frac{B_g -B_{gi} + B_o(R_{vi} - R_v) + R_s(B_{gi}R_v - B_gR_{vi})}{1-R_sR_v}\bigg] -W_pB_w + W_{inj}B_{winj} + G_{inj}B_{ginj} \,+\, \\
(N_{foi}B_{oi} + G_{fgi}B_{gi})\frac{c_f + c_wS_{wi}}{1- S_{wi}} \Delta p + W_e \, + 
\sum_{j=1}^{Neigh} \sum_{\alpha\,=\,o,g,w} \int_0^t \frac{T_{j}}{\bar{\mu}^t_{\alpha}}k_{r\alpha}(S^t)(p_{j,\alpha}^t - p_{\alpha}^t - \bar{\rho}_\alpha^tg(z_j - z))\,dt = 0\,.
\end{gathered}
\ee
As the second equation, we consider the flowing bottomhole pressure (FBHP) as the operating parameter. In this case the second equation describes the inflow performance relation (IPR). This equation relates the total liquid production rate with the flowing bottomhole pressure and the average reservoir pressure at each time step. Here, we consider the Vogel's equation \citep{Tarek} as:
\be \label{Eq_forecast2}
Q_L = Q_{L,max}\,\Big[ 1 - 0.2\frac{p_{wf}}{p} - 0.8(\frac{p_{wf}}{p})^2 \Big]\,{N_{producers}} = \frac{\Delta W_p + \Delta N_p}{\Delta t}\,,
\ee
where $Q_L$ is the total liquid (oil and water) production rate and $Q_{L,max}$ is the maximum liquid flow rate at zero FBHP, also called absolute open flow (AOF). $P_{wf}$ is the flowing BHP for an average well, $p$ is the average reservoir pressure, $N_{producers}$ is the number of active producers at each time step, and $\Delta W_p = W^{n+1}_p - W^{n}_p$ and $\Delta N_p = N^{n+1}_p - N^{n}_p$ are the cumulative volume of water and oil production during one time step respectively. \\
In the third equation, the relationship between field measurements of water/oil ratio ($WOR$) and fluid and rock-fluid characteristics including $PVT$, viscosity, and relative permeability information will be considered as:
\be \label{Eq_forecast3}
WOR = \frac{k_{rw}(S_o,S_w)}{k_{ro}(S_o,S_w)} \times  \frac{\mu_o(p) B_o(p)}{\mu_w(p) B_w(p)} = \frac{\Delta W_p}{\Delta N_p}\,.
\ee
The fourth equation relates the field measurements of gas/oil ratio (GOR) to characteristic properties of oil, gas, and rock including viscosity, PVT, and relative permeability as:
\be \label{Eq_forecast4}
GOR = R_s(p) + \frac{k_{rg}(S_o,S_w)}{k_{ro}(S_o,S_w)} \times \frac{\mu_o(p)B_o(p)}{\mu_g(p)B_g(p)} = \frac{\Delta G_p}{\Delta N_p}\,.
\ee
In Equations \mref{Eq_forecast3, Eq_forecast4} above, $k_{rg}$, $k_{ro}$, and $k_{rw}$ are the relative permeabilities of rock to gas, oil, and water in the presence of the three phases respectively, and $\mu_o$, $\mu_g$, and $\mu_w$, respectively represent the viscosity of oil, gas, and water ($cP$) at the current reservoir pressure. Likewise, $B_o$, $B_g$, and $B_w$ correspond to the formation volume factors (RB/STB) of oil, gas, and water at the current reservoir pressure respectively.\\
It is worth mentioning that the relative permeability ratios in both Equations \mref{Eq_forecast3, Eq_forecast4} are different from the ones measured in the lab using cores. The former also depends on the development and operation history of the field, whereas the latter only depends on the rock and fluid properties and their interaction. \\
In order to solve the Equations \mref{Eq_forecast1,Eq_forecast2,Eq_forecast3,Eq_forecast4} as a non-linear system of equations, we write them in the linear Residual form and after building the Jacobian matrix, we solve for the unknowns vector ($\vec{X}$)  as:
\bsa 
 \label{Eq_Res2}
\vec{x}_b^{\,n} = [x_{1}, x_{2}, x_{3}, x_{4}]^{\,n}_b =  [{p}, {N}_{p}, {G}_{p}, {W}_{p}]^{\,n}_b \quad \text{with block index:~ } b = 1, ...\,, N\,, \\
\vec{X}^{\,n} = [\vec{x}_1^{\,n}, ..., \vec{x}_N^{\,n}]\,,\\
\vec{R}(\vec{X}^{\,n}) = \vec{0}\,,\\
\label{Eq_Jac2}
\tensor{J}\vec{\delta_X} = -\vec{R} \quad \text{where~~} \tensor{J} = \frac{\partial \vec{R}}{\partial \vec{X}}\,,\\
\vec{X}^{\,n+1} = \vec{X}^n + \vec{\delta_X}\,.
\esa
Here, we build the unknowns vector over all the four unknowns (i.e. $p, N_p, G_p, W_p$) and over all the blocks (i.e. $b= 1, ..., N$) in a block-based order (i.e. $\vec{X} = \Big[(p, N_p, G_p, W_p)_1\,, ...,\,(p, N_p, G_p, W_p)_N\Big]$). Ordering the unknowns this way leads to a less sparse block-diagonal Jacobian matrix and leads to a smoother convergence of the linear solver once solving Equation \eqref{Eq_Jac2} to find $\vec{\delta_X}$. The Jacobian matrix has $4\times4$ diagonal blocks (local Jacobian elements) and some non-diagonal non-zero elements due to non-local connectivity of the blocks in the reservoir and pressure derivatives with respect to the adjacent blocks. \\
First, we present the Equations \mref{Eq_forecast1,Eq_forecast2,Eq_forecast3,Eq_forecast4} in a Residual form. The Residual form of Equation \eqref{Eq_forecast1} is similar to Equations \mref{Eq_ResLoc1, Eq_ResLoc2} presented before. The Residual form of Equation \eqref{Eq_forecast2} would be:
\be
R_2 = (W_p^{n+1} - W_p^n) + (N_p^{n+1}-N_P^n) - \Big[ 1 - 0.2\,(\frac{p_{wf}}{p}) - 0.8\,(\frac{p_{wf}}{p})^2 \Big]\,{\Delta t\, Q_{L,max}N_{producers}}\, .
\ee
The Residual form of Equation \eqref{Eq_forecast3} can be written as:
\be
R_3 = \frac{k_{rw}(S^{n+1}_o,S^{n+1}_w)}{k_{ro}(S^{n+1}_o,S^{n+1}_w)}  \frac{\mu_o(p^{n+1}) B_o(p^{n+1})}{\mu_w(p^{n+1}) B_w(p^{n+1})}(N_p^{n+1}-N_P^n) - (W_p^{n+1} - W_P^n)\,.
\ee
And finally, the Residual form of the Equation \eqref{Eq_forecast4} becomes: 
\be
R_4 = R_s(p^{n+1}) + \frac{k_{rg}(S^{n+1}_o,S^{n+1}_w)}{k_{ro}(S^{n+1}_o,S^{n+1}_w)}  \frac{\mu_o(p^{n+1})B_o(p^{n+1})}{\mu_g(p^{n+1})B_g(p^{n+1})}(N_p^{n+1}-N_P^n) - (W_p^{n+1} - W_P^n)\,.
\ee
The corresponding Jacobian equations for the Residual terms $R_2, R_3, R_4$ can be derived in a similar way as Equation \eqref{Eq_Jacob1}.
\subsection{Global History Matching.} \label{sec_history_matching}
We use an efficient hierarchical history matching approach to update the model parameters. Our history matching approach is a global history matching where large-scale features such as transmissibility between the blocks, aquifer size and strength, average properties of each block, and original oil in plae per block
are adjusted to match fluid production and bottomhole pressure data using \textit{ensemble smoother robust Levenberg-Marquardt (ES-rLM)} \citep{Bi2017} history matching workflow. Rather than a long process of finding the minimum objective function value using for instance a genetic algorithm, the ES-rLM is a Bayesian approach that accounts for probabilistic characteristics, which requires much fewer simulation runs. \citet{Salehi2019a} have recently applied this algorithm for solving a full-physics history matching probelm.

Although the ES-rLM algorithms have been discussed elsewhere \citep{Bi2017}, we provide a brief review of the main equations here for completeness. The ensemble equation derived based on the Levenberg-Marquardt algorithm can be written as:
\be
m^{new}_j = m_j^i + C_{MD}^i (C_{DD}^i + \alpha^iC^i_{D})^{-1}(d_{i,j} - g(m_j^i))\,,
\ee
where $m_j^i$ is an ensemble, $j=1,..., N_e$ at the $i^{th}$ iteration. The relationship between the ensembles and the model parameters can be described as:
\be
C_{MD}^i=\frac{1}{(N_e-1)} \sum_{j=1}^{N_e}(m_j^i - \overline{m^i}) [g(m_j^i)- \overline{g(m_j^i)}]^T\,,
\ee 
where $g(m)$ is the linearized forward model around the ensemble mean, and the mean of the predicated data is:
\be
\overline{g(m_j^i)} = \frac{1}{N_e} \sum_{j=1}^{N_e} g(m_j^i)\,,
\ee
and,
\be
C_{DD}^i = \frac{1}{(N_e - 1)}\sum_{j=1}^{N_e} [g(m_j^i)- \overline{g(m_j^i)}][g(m_j^i)- \overline{g(m_j^i)}]^T\,.
\ee
To account for cases that the data covariance matrix is not a diagonal matrix and doesn't follow a Gaussian distribution, we apply an extension to the robust ensemble smoother (rES-LM):
\be
C_D^i = diag(\frac{1}{w_1}, ..., \frac{1}{w_{N_d}})\,.
\ee 
Here, \textit{Cauchy weight function} is used for the measurement error as:
\be
w_k = \frac{1}{1+r_k^2}\,, ~~~ \text{with }k=1, ..., N_d\,,
\ee
where $r$ is the vector of Residuals from the previous iteration divided by the estimate of the standard deviation of the Residual.
\section{Results and Discussion} \label{sec_results}
\subsection{Case Study Reservoir.}
\begin{figure}[!ht]
\centering
\includegraphics[width=0.35\textwidth]{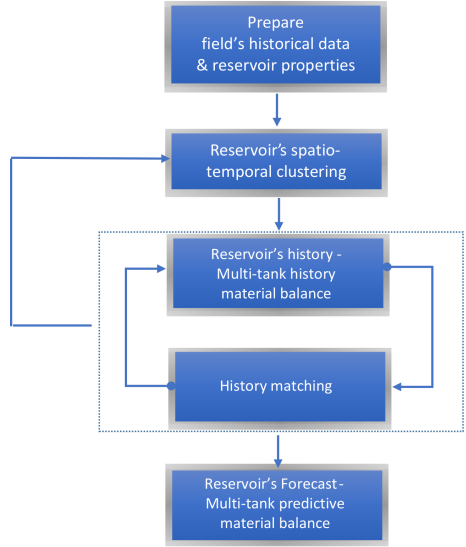}
	\caption{Flowchart showing the order of workflow steps from raw reservoir's input data to reservoir history matching and forecast}
	\label{fig_flowchart}
\end{figure}

In this section, we apply our modeling framework described in the former sections to a real-world reservoir and the procedures follow the steps presented in Figure \ref{fig_flowchart}. The reservoir considered in this work is a mature field with more than 40 years of recovery. The field has 404 wells with 340 being producers and 64 being injectors. The production from the reservoir includes both gas and oil and the injections include both water and gas. Due to the data sharing policy, we are unable to present more details than what is shown in the following figures and charts and we have masked/normalized the quantitative information.

\subsection{Spatio-temporal Clustering.}
\begin{figure}[!ht]
\centering
        \begin{subfigure}[b]{0.2\textwidth}
            \includegraphics[width=1\linewidth]{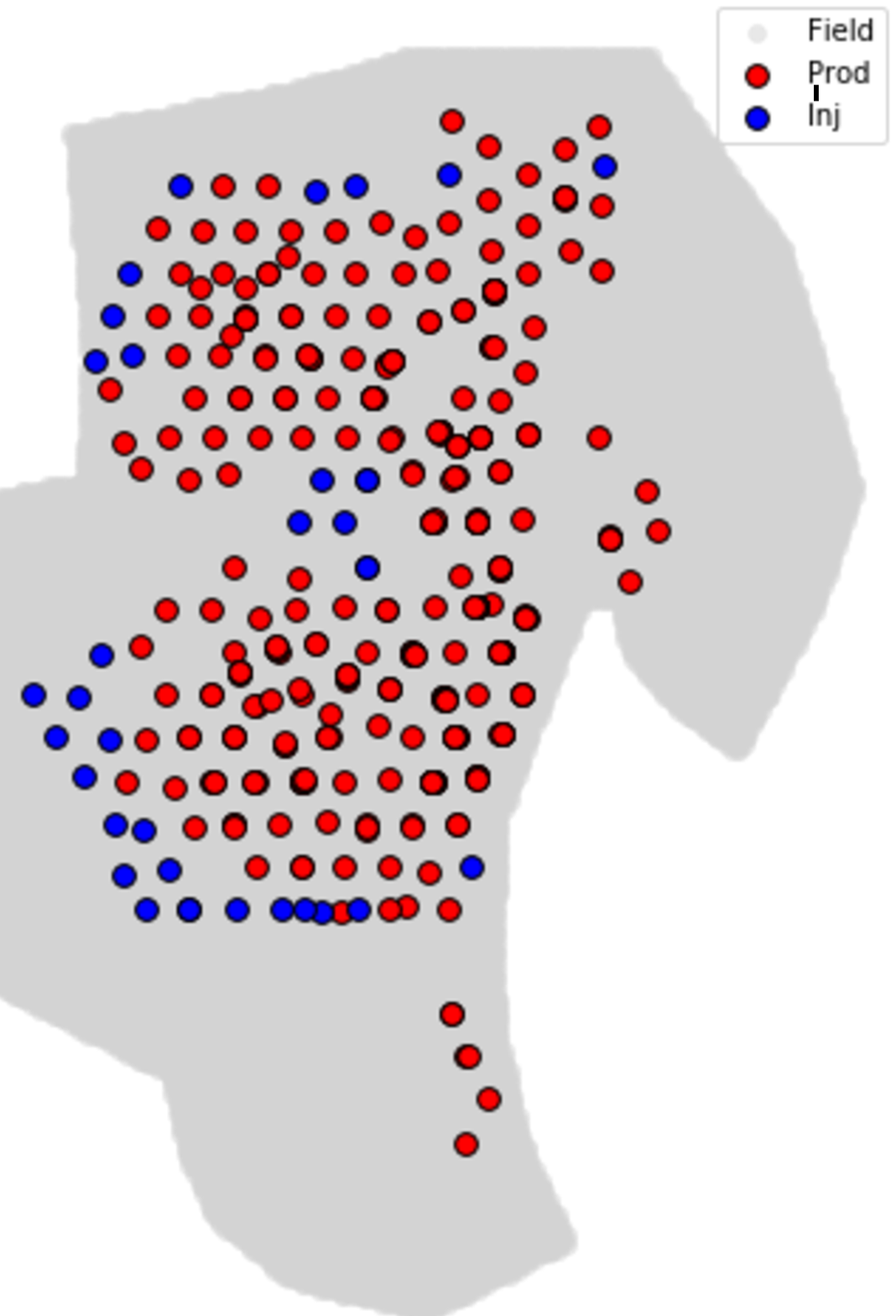}
            \caption{}
            \label{fig_wells_layout}
        \end{subfigure}%
                \begin{subfigure}[b]{0.2\textwidth}
                \includegraphics[width=1\linewidth]{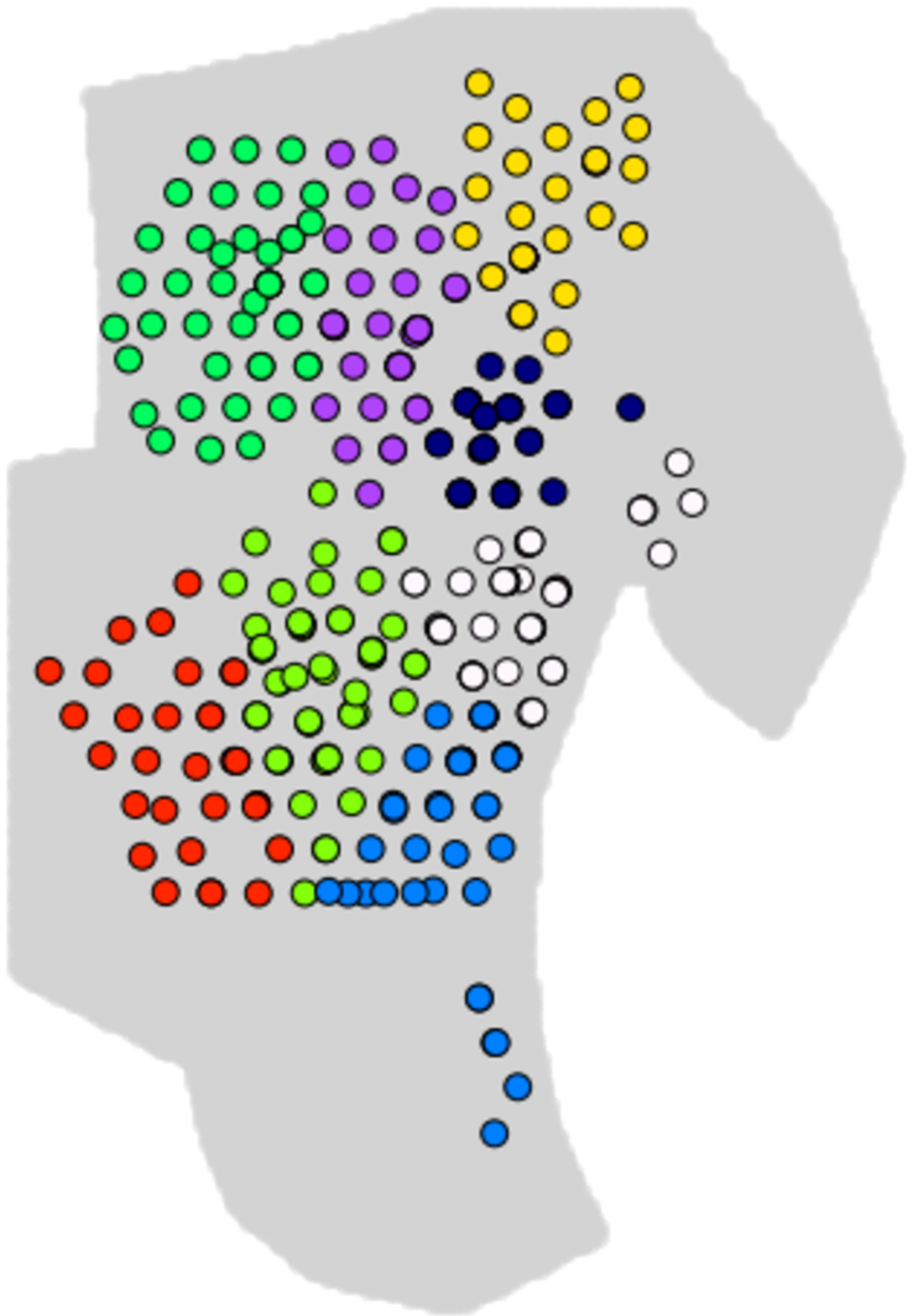}
                \caption{}
                \label{fig_wells_clustered}
        \end{subfigure}
        \begin{subfigure}[b]{0.2\textwidth}
                \includegraphics[width=1\linewidth]{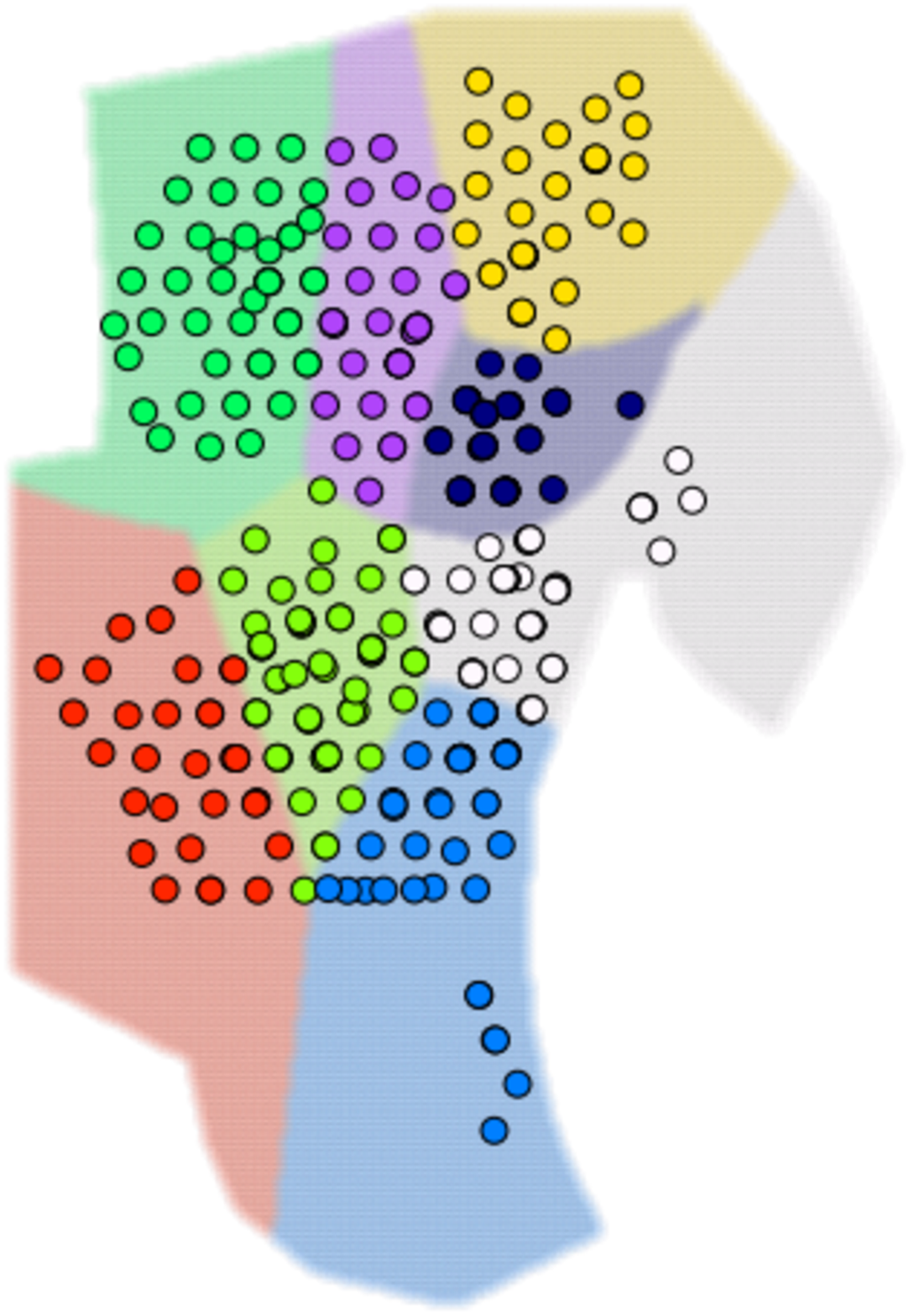}
                \caption{}
                \label{fig_zones_clustered}
        \end{subfigure}%
        \caption{(a). Producer wells (in red) and injector wells (in blue) layout; (b). well-level cluster illustration; (c). hyperplane identification of well-level clusters using support vector machine}\label{fig_3ClusterPics}
\end{figure}

In this part, we first present step-by-step our clustering framework presented before. We consider both static and temporal features of both categorical and numerical data types. We consider well types, wellhead locations, the number of perforations, history of gas, oil, and water production, history of gas and water injection, and PVT measurements. Also, we take into account geological features by assigning labels to each well for types of formation and for sealing faults.
\begin{figure}[!ht]
\centering
\includegraphics[width=0.8\textwidth]{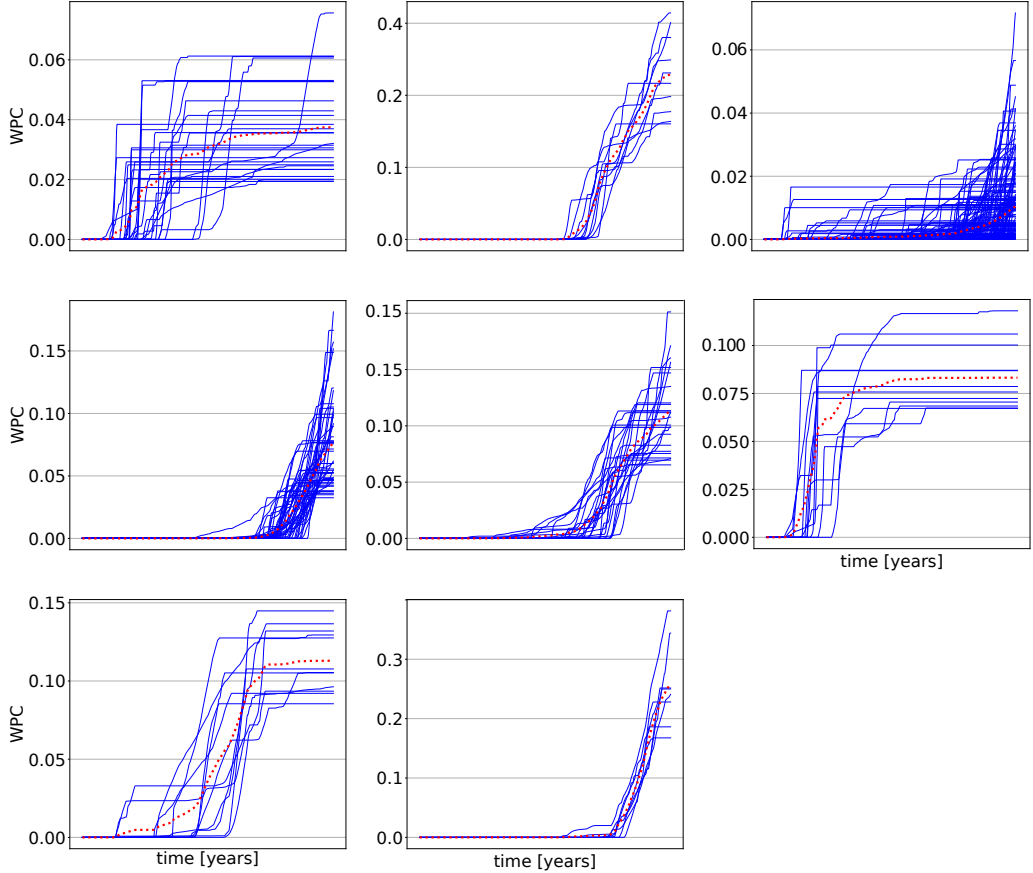}
	\caption{Temporal clustering of production wells for cumulative water production (WPC)}
	\label{fig_temporal_clustering}
\end{figure}
For illustration purposes, in Figure \ref{fig_temporal_clustering}, we are showing an example of the adaptive temporal clustering framework presented before for the cumulative water production (WPC) of the producer wells, where prior to temporal clustering the WPC is normalized and an internal variation threshold is specified and tuned for adaptively finding the proper number of clusters. As Figure \ref{fig_temporal_clustering} shows a great separation of scales and behavior is achieved by the temporal clustering and the time series in each cluster are similar in terms of the range of values and their overall transient behavior.\\
Furthermore, for illustration purposes, in Figure \ref{fig_3ClusterPics} we illustrate how from the well-level clusters, we compartmentalize the reservoir at the field-scale in a spatially constrained approach. Figure \ref{fig_wells_layout} shows the layout of producer and injector wellheads in the reservoir. Considering the above mentioned static and temporal categorical and numerical features and choosing a high number of clusters to stress-check the clustering robustness and clusters coherency (i.e. eight in this case), the well-level clusters are shown in Figure \ref{fig_wells_clustered}. As described in the clustering part of this manuscript, using the labeled clustered wells in Figure \ref{fig_zones_clustered}, we use the SVM approach and find the separating hyperplanes of clustered wells in the physical space.

\begin{figure}[!ht]
\centering
\includegraphics[width=0.2\textwidth]{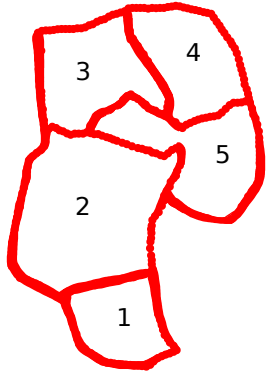}
	\caption{Spatio-temporal clustered reservoir - $k=5$ as the number of clusters}
	\label{fig_final_clusters}
\end{figure}

Now, in order to illustrate the proposed approach for a compartmentalized reservoir modeling, we use the clustered reservoir with five clusters, where $k=5$ as the number of clusters has been found using the elbow criterion and convergence of the loss functions in the clustering framework. Figure \ref{fig_final_clusters} shows the layout of 5-cluster reservoir where the zones that are connected to each other are considered to have multiphase flow communications governed by pre-assigned transmissibility values within a range of uncertainty derived from geological maps, which later on will be history matched. 
\subsection{Multi-Tank Reservoir's History Material Balance Results.}
\begin{figure}[!ht]
\centering
\includegraphics[width=0.8\textwidth]{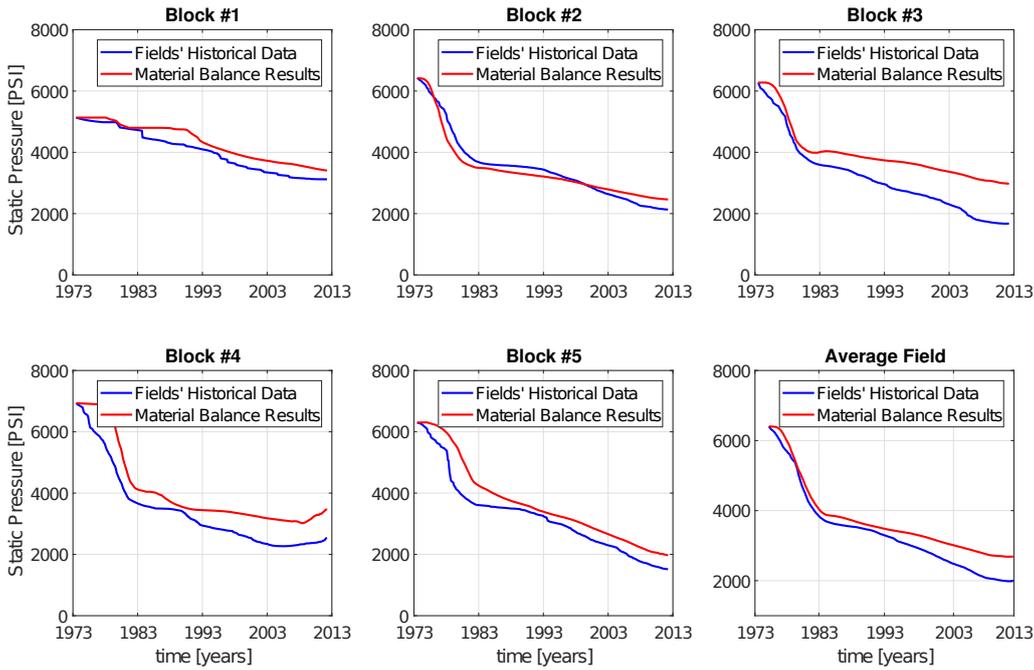}
	\caption{Material balance results vs. field's actual historical data for the 5-compartments clustered reservoir of Figure \ref{fig_final_clusters}}
	\label{fig_history_results}
\end{figure}
In this section, we present the multi-tank history material balance results for the 5-compartment clustered reservoir shown in Figure \ref{fig_final_clusters}. We calculate average properties of each block, e.g., porosity, initial saturations, formation compressibilities, etc., from the available data of the field and the reports by geologists. Figure \ref{fig_history_results} shows the comparison of the multi-tank material balance results for the average pressure of each compartment compared with the field's actual historical average pressure data in each block.

\subsection{Multi-Tank Reservoir's History Matching Results.}
\begin{figure}[!ht]
\centering
\includegraphics[width=0.8\textwidth]{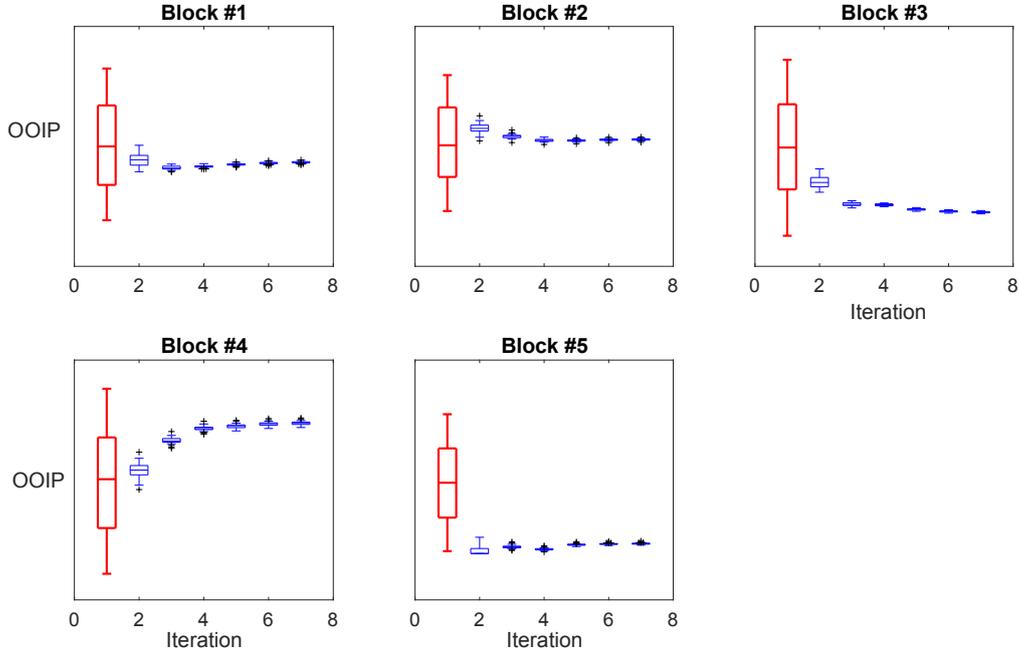}
	\caption{Box plots for the original oil in place as the function of number of iterations during the history matching process for the 5-compartment clustered reservoir of Figure \ref{fig_final_clusters}}
	\label{fig_historymatch_boxplots}
\end{figure}

As there are uncertainties attributed to the inputs of the material balance approach, using the available production data, injection data, reservoir's pressure, and etc., we carry out the global history matching explained before. We consider as the uncertain parameters of history matching the maximum volume of water encroachable from aquifer in each block ($W_{ei}~[RB]$), Fetkovich aquifer transmissibility ($J~[RB/(psi.D)]$), the transmissibility between connected blocks ($T_{ij}~[mD.ft]$), the maximum transmissibility used for scaling ($T_{max}~[mD.ft]$), and most importantly the initial oil in place for each compartment ($OOIP~[STB]$). We set as the objective function of history matching, the overall mean squared error of the pressure mismatch in each compartment between the multi-tank material balance and the actual field's data. We consider 5 to 10 number of ensembles each with 50 to 100 ensemble members. We set 0.1\% to 1\% as the threshold of change for both objective function value and uncertain parameters value between two consecutive iterations as the termination criterion. Also, we consider the ratio of the initial measurements error to the maximum observation to be less than 1\%. \\
Figure \ref{fig_historymatch_boxplots} shows the box plots for the original oil in place value in each of the five compartments in the clustered reservoir in Figure \ref{fig_final_clusters} versus the number of iterations during the history matching process. As observed in Figure \ref{fig_historymatch_boxplots}, by starting with a range of uncertainty in each block, as history matching progresses for a few iterations, the original oil in place values all converge.\\

\begin{figure}[!ht]
\centering
\includegraphics[width=0.8\textwidth]{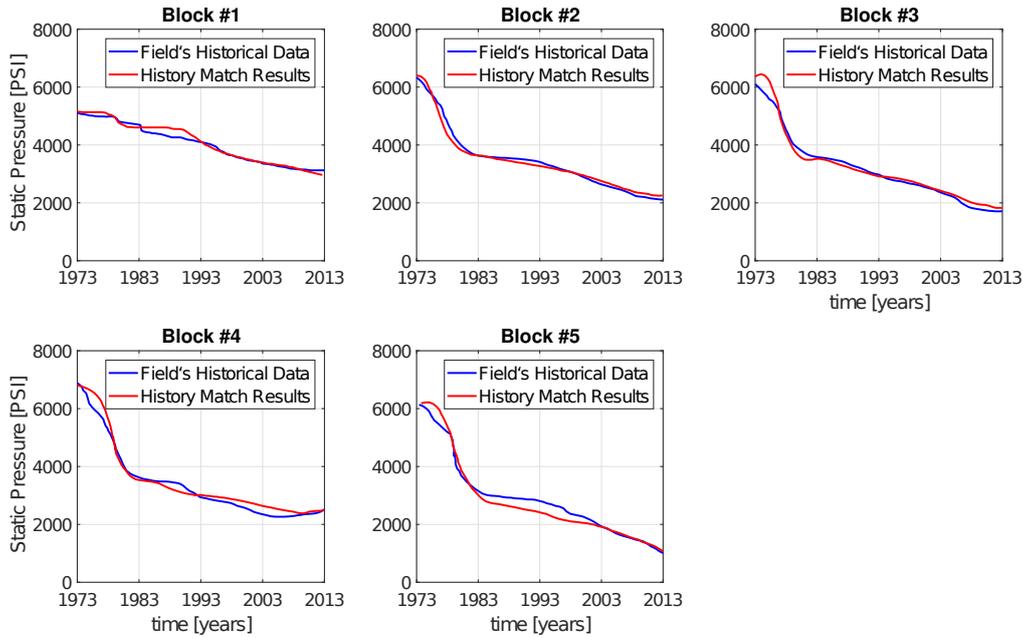}
	\caption{History matched material balance results vs. field's actual historical data for the 5-compartment clustered reservoir of Figure \ref{fig_final_clusters}}
	\label{fig_historymatch_results}
\end{figure}

Figure \ref{fig_historymatch_results} shows the comparison of pressure solutions from the multi-tank material balance approach after history matching the uncertain parameters (i.e. $W_{ei}, J, T_{ij}, T_{max}, OOIP$) with the field's historical average pressure in each block. Comparing pressure solutions in Figure \ref{fig_historymatch_results} with the ones in Figure \ref{fig_history_results}, one can see considerable improvements gained through history matching the uncertain input parameters in the multi-tank material balance approach, which have led to a decrease in the overall pressure mismatch by one order of magnitude.

\subsection{Multi-Tank Reservoir's Predictive Material Balance Results.}
\begin{figure}[!ht]
\centering
                \begin{subfigure}[c]{0.5\textwidth}
                \includegraphics[width=7cm,height=5.7cm]{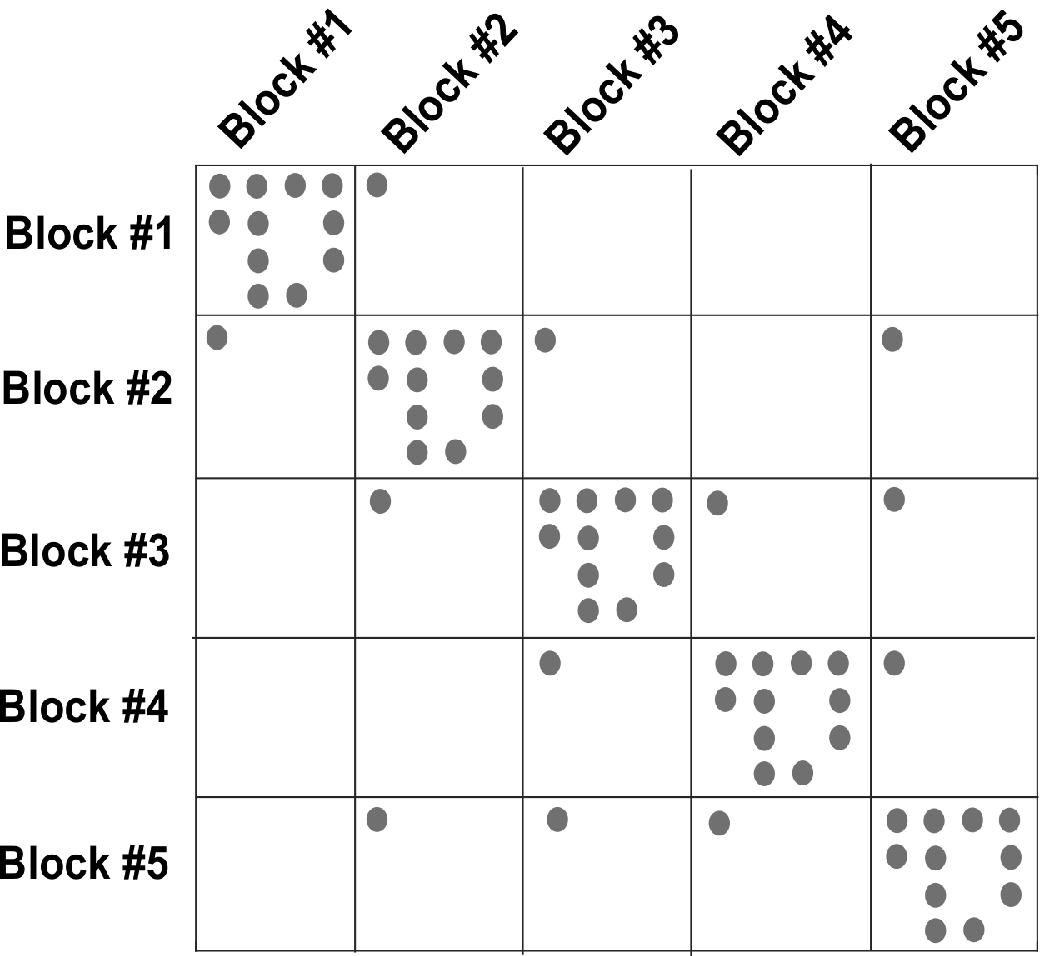}
                \vspace{20pt}
                \caption{}
                \label{fig_jacobian}
        \end{subfigure}%
                \begin{subfigure}[c]{0.5\textwidth}
                \includegraphics[width=8cm,height=8cm,keepaspectratio]{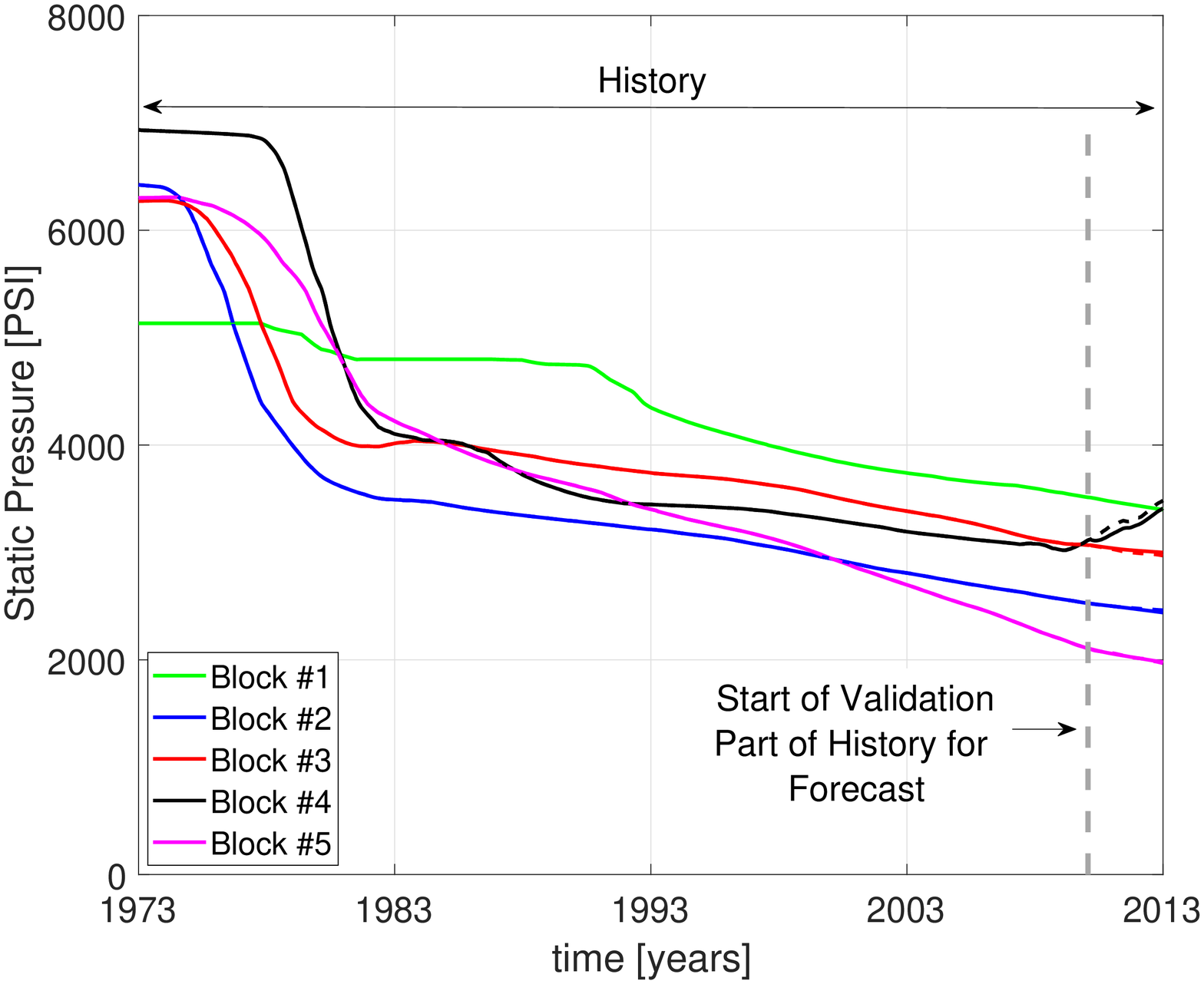}
                \caption{}
                \label{fig_forecastResults}
        \end{subfigure}
        \caption{(a). Jacobian matrix structure for the 5-compartment clustered reservoir of Figure \ref{fig_final_clusters} for multi-tank predictive material balance - gray circles represent the non-zero elements in the matrix. (b). Validation of the forecast approach - last 10\% of the historical data is masked and compared with the forecast pressure solution for the 5-compartment clustered reservoir of Figure \ref{fig_final_clusters} - the vertical dashed line shows the start of forecast -  Dashed lines show the actual history and solid lines show the forecast results}\label{fig_forecastandJacobian}
\end{figure}

After history matching the uncertain parameters of the multi-tank history material balance, we perform forecast based on the future recovery plans for the reservoir. In order to illustrate the robustness of our forecast approach, we carry out a validation test. For the validation test, we use the historical data that is available from the reservoir and perform a \textit{blind test} analysis where we mask the last 10\% of the history (validation's forecast region) and assume that only the first 90\% of the history (validation's history region) is known. Then, knowing the reservoir parameters in the last time step of the validation-history region, we perform a forecast with the known recovery plan of the reservoir. \\
For the forecast's validation test, we use the same 5-compartment clustered reservoir shown in Figure \ref{fig_final_clusters}. The Jacobian matrix formed for solving the forecast set of equations (Equation \eqref{Eq_Jac2}) is shown in Figure \ref{fig_jacobian}. As shown, the Jacobian matrix is dominantly block diagonal with a few non-zero off-diagonal terms due to the non-zero pressure derivatives with respect to neighboring blocks. For instance, as the block (5) is connected to blocks (2), (3), and (4) in Figure \ref{fig_final_clusters}, in the Jacobian matrix, non-zero elements are present in the 5th row's columns 2, 3, and 4. As also mentioned before, this compactness of the Jacobian matrix is achieved due to the presented optimal ordering of unknowns in Equation \eqref{Eq_Res2}.\\
Figure \ref{fig_forecastResults} shows the results for the forecast validation test where the last 10\% of the historical data is masked and compared with the forecast pressure solution for the 5-compartment clustered reservoir of Figure \ref{fig_final_clusters}. The vertical dashed line shows the starting point of the forecast and the colored dashed lines show the actual history and solid lines show the forecast results. The great agreement of the forecast results in the validation part of Figure \ref{fig_forecastResults} with the field data shows the robustness and accuracy of our proposed forecast framework.

\section{Summary and Conclusions}
In this work, we represented the reservoir as a network of discrete compartments with neighbor and non-neighbor connections for a fast, yet accurate modeling of oil and gas reservoirs. A key element of such high-level reservoir analysis is the automatic and rapid detection of coarse-scale compartments with distinct static and dynamic properties. We presented a hybrid framework specific to reservoir analysis for an automatic detection of clusters in space using spatial and temporal field data, coupled with a physics-based multiscale modeling approach. A non-local formulation for flow in porous media was presented, in which the reservoir is represented by an adjacency matrix describing the neighbor and non-neighbor connections of comprising compartments. We clustered the reservoir into distinct compartments, in which the direction-dependent multiphase flow communication is a function of non-local phase potential differences. The proposed framework was successfully applied to a major field with a couple of hundreds of wells and a long production history. Leveraging the fast forward model, an efficient ensemble-based history matching framework was applied to reduce the uncertainty of the global reservoir parameters, such as inter-blocks and aquifer-reservoir communications, fault transmissibilities, and oil in place values per block. The ensemble of history matched models were then used to provide a forecast for different field development scenarios.

In summary, a novel hybrid approach was presented in which we couple a physics-based non-local modeling framework with data-driven clustering techniques to provide a fast and accurate multiscale modeling of compartmentalized reservoirs. This work also presented a comprehensive workflow on spatio-temporal clustering for reservoir studies applications that well considers the clustering complexities, the intrinsic sparse and noisy nature of the data, and the interpretability of the outcome. The proposed framework can also be applied to fractured reservoirs as a special case of multi-tank material balance formulation, where multiple tanks can account for matrix and fracture components with arbitrary connectivity configuration. Furthermore, the proposed framework can be applied to three-dimensional network models where gravity effects are significant, accounting for both pressure-driven and gravity-driven flows.


\section{Nomenclature}
\begin{center}
\footnotesize
\begin{longtable}{ | m{2.5 em}  m{7cm} m{4.2em} m{7cm} |} 
\hline
\vspace{5pt}
 &  &   &    \\ 
$B_g$ & Gas formation volume factor &  $B_{gi}$  &  Initial gas formation volume factor  \\ 
$B_{ginj}$ & Injection gas formation volume factor &  $B_o$ & Oil formation volume factor   \\ 
$B_{oi}$ & Initial oil formation volume factor &  $B_w$ & Water formation volume factor  \\ 
$B_{winj}$ & Injection water formation volume factor &  $c_f$ & Formation compressibility  \\ 
$c_w$ & Water compressibility & $J$ & Fetkovich aquifer transmissibility  \\ 
$G$ & Initial gas in place & $G_{inj}$ & Cumulative volume of gas injected\\ 
$G_{fg}$ & Volume of gas component in the gas phase & $G_{fgi}$ & Initial volume of gas component in the gas phase\\
$G_p$ & Cumulative volume of gas produced & $N$ & Initial oil in place \\ 
$N_{fo}$ & Volume of oil component in the oil phase & $N_{foi}$ &Initial volume of oil component in the oil phase\\ 
$N_p$ & Cumulative volume of oil produced & $p$ &  Reservoir pressure\\ 
$p_i$ & Initial reservoir pressure & $\Delta p$ & Pressure depletion  \\ 
$R_s$ & Solution-gas to oil ratio &  $R_{si}$ & Initial solution-gas to oil ratio \\ 
$R_v$ & Volatile-oil to gas ratio & $R_{vi}$ &  Initial volatile-oil to gas ratio\\
$S_{wi}$ & Initial water saturation & $t$ & time\\
$\theta$ & Aquifer encroachment angle & $V_{\phi}$ & Reservoir effective pore volume \\
$W$ & Initial water in place & $W_e$ & Aquifer's cumulative encroached water volume\\
$Q_L$ & Total liquid (oil and water) production rate & $Q_{L,max}$ & Maximum liquid flow rate at
zero FBHP\\
$W_{ei}$ & Aquifer's maximum encroached water volume & $W_{inj}$ & Cumulative volume of water injected\\	
$W_p$ & Cumulative volume of water produced & $\alpha$ & Phase index (oil, water or gas) \\
$\bar{\mu}_\alpha$ & Viscosity of phase $\alpha$ & $k_{r\alpha}$ & Relative permeability of phase $\alpha$  \\
$T_{ij}$ & Transmissibility of block $i$ and $j$  & $V_b$ & Reservoir bulk volume  \\
$n$ & Time step index & $K$ & Absolute permeability \\
$\tensor{J}$ & Jacobian matrix & $\vec{R} $ & Residual vector \\
$DTW$ & Dynamic Time Warping  & $SVM$ & Support Vector Machine  \\
$KL$ & Kullback-Liebler   & $ES$-$rLM$ & Ensemble smoother
robust Levenberg-Marquardt   \\
$AOF$ & Absolute open flow   & $GTTI$ & Generalized travel-time inversion   \\
$FBHP$ & Flowing bottomhole pressure   & $WOR$ & water/oil ratio   \\
$IPR$ & Inflow performance relation  & $ROM$ & Reduced order modeling   \\
$p_{wf}$ & Flowing bottomhole pressure   & $GOR$ & gas/oil ratio   \\
$N,\,M$ & Sequence length in DTW & $C$ & Cost function in DTW \\
$\zeta$ & Cost function in k-prototypes & $D$ & Distance function in k-prototypes \\
$\delta$ & Weight for categorical attributes in k-prototypes & $C$ & Mode in k-prototypes \\
\hline
\end{longtable}
\end{center}

\bibliographystyle{myplainnat}

\end{document}